\documentclass[10pt,conference]{IEEEtran}
\IEEEoverridecommandlockouts
\usepackage{cite}
\usepackage{amsmath,amssymb,amsfonts}
\usepackage{algorithmic}
\usepackage{graphicx}
\usepackage{textcomp}
\usepackage{xcolor}
\usepackage{multirow}
\usepackage{makecell}
\usepackage{float}
\usepackage{balance}
\usepackage{algorithm}
\usepackage{array, caption, threeparttable}
\begin{document}

\title{Representation Learning of Knowledge Graph for Wireless Communication Networks
}

\author{\IEEEauthorblockN{Shiwen~He\IEEEauthorrefmark{1}\IEEEauthorrefmark{2}\IEEEauthorrefmark{3},
		Yeyu~Ou\IEEEauthorrefmark{1},
		Liangpeng~Wang\IEEEauthorrefmark{2},
		Hang~Zhan\IEEEauthorrefmark{2},
		Peng~Ren\IEEEauthorrefmark{2},
		Yongming~Huang~\IEEEauthorrefmark{2}\IEEEauthorrefmark{3}}
	
	\IEEEauthorblockA{\IEEEauthorrefmark{1}\emph{The School of Computer Science and Engineering, Central South University, Changsha 410083, China.}}
	\IEEEauthorblockA{\IEEEauthorrefmark{2}\emph{The Purple Mountain Laboratories, Nanjing 211111, China. }}	
	\IEEEauthorblockA{\IEEEauthorrefmark{3}\emph{The National Mobile Communications Research Laboratory, Southeast University, Nanjing 210096, China.}}
	Email: \{shiwen.he.hn, ouyeyu\}@csu.edu.cn, \{wangliangpeng, zhanxing, renpeng\}@pmlabs.com.cn, huangym@seu.edu.cn
}

\maketitle

\begin{abstract}
With the application of the fifth-generation wireless communication technologies, more smart terminals are being used and generating huge amounts of data, which has prompted extensive research on how to handle and utilize these wireless data. Researchers currently focus on the research on the upper-layer application data or studying the intelligent transmission methods concerning a specific problem based on a large amount of data generated by the Monte Carlo simulations. This article aims to understand the endogenous relationship of wireless data by constructing a knowledge graph according to the wireless communication protocols, and domain expert knowledge and further investigating the wireless endogenous intelligence.
We firstly construct a knowledge graph of the endogenous factors of wireless core network data collected via a 5G/B5G testing network. Then, a novel model based on graph convolutional neural networks is designed to learn the representation of the graph, which is used to classify graph nodes and simulate the relation prediction. The proposed model realizes the automatic nodes classification and network anomaly cause tracing. It is also applied to the public datasets in an unsupervised manner. Finally, the results show that the classification accuracy of the proposed model is better than the existing unsupervised graph neural network models, such as VGAE and ARVGE.
\end{abstract}


\section{Introduction}

The gradual popularization and large-scale deployment of fifth-generation (5G) mobile networks have realized a leap forward in communication capability and service quality for public users. Meanwhile, a large number of smart terminals are widely used, which makes the structure of wireless communication networks more complex and poses a significant challenge to the traditional model-based algorithms for designing and optimizing wireless networks. On the other hand, many access terminals generate a huge amount of wireless data, so data-driven intelligent learning brings new ideas for optimizing communication networks. However, a large portion of the massive data generated by wireless communication networks is the underlying data, which is difficult to be used directly in artificial intelligence (AI) algorithms.
%In order to better optimize the network by using wireless communication network data for artificial intelligence, it is necessary to sort out the correlation of data and understand the meaning of the data.

Currently, studies on intelligent wireless communications focused on optimizing the performance criteria of wireless networks, such as channel selection\cite{37}, cache optimization\cite{38}, beam selection\cite{41}, etc, using wireless data generated by Monte Carlo simulations. In addition, many studies focus on the upper-layer data above the network layer, such as unmanned delivery. However, these studies do not reveal the meaning of the wireless communication networks' data fields and do not analyze in-depth the relationships between the data. Meanwhile, the lack of a publicly available and trusted database of wireless data makes most of the existing algorithms use simulated data directly instead of the real data to evaluate the performance.
%As a result, there is a gap in the study of the underlying data of wireless communication networks. Furthermore, the real underlying data is difficult to be used, which will bring difficulties for applying the AI algorithms to real communication networks in the future.

On the other hand, the challenge in studying the underlying data of communication networks is that these data are difficult to understand and access directly\cite{35}, requiring one to understand the content of the data fields and the relationships between the individual fields based on the wireless communication protocols. Due to the diverse contents and complex relationships involved in wireless communication protocols, non-experts and even experts are under tremendous pressure to fully understand and utilize this knowledge. There is an urgent need for a way to combine wireless communication knowledge and wireless data to help people make sense of the relationships between the data fields.
%We urgently need a method that combines wireless communication knowledge and wireless communication network data, and uses the knowledge to represent the data visually to help sort out the relationships between the data.

Among various data representation methods, knowledge graph (KG) using both knowledge rules and data awareness is an effective way to transform data into usable knowledge and to update data dynamically\cite{1}. Due to its powerful relational representation and interpretability, KG has been used in many fields, such as medicine, finance, etc. Introducing KG into wireless data can help people understand the meaning of data fields through knowledge visualization of obscure underlying data and mine the relationships between data fields and sort out the correlations between data for downstream network optimization tasks. In addition, KG is a tool to make complex and diverse wireless data structured, which benefits easy-to-use data for wireless communication optimization problems.

Inspired by the vacancy of wireless data knowledge graph and the wide application of knowledge graphs in vertical domains, we construct a KG of wireless data of the core network data and propose a novel knowledge representation learning (KRL) model for learning and mining knowledge graphs. We construct graph autoencoders based on graph convolutional neural networks (GCN) to learn the knowledge representation of KG by combining with generative adversarial networks (GAN)\cite{18} to enhance the generative power. To compare the model's performance with other typical unsupervised graph neural networks, such as VGAE, ARVGE, etc., we perform an unsupervised node classification task on a public dataset. The simulation results show that the proposed model exhibits superior performance in terms of classification accuracy compared to other models. On the other hand, the proposed model is used to learn the representation learning matrix and node link probability matrix of the KG of the core network data, which are used for node classification and relation prediction of wireless data. The results show that the model overlaps the classification results of data fields and manual classification results and exhibits a high accuracy of relation prediction between data fields. Finally, based on the high relation prediction rate, we propose a scheme to use the relation prediction results for downstream network anomaly cause tracing task.

%The main contribution of this paper is to construct a novel graph representation learning framework to learn and mine knowledge graphs of wireless communication networks. The main contributions are as follows:
%
%\begin{itemize}
%\item In order to address the current gap in the study of complex and difficult to understand underlying wireless communication data, we introduce knowledge graph technology to build a KG of wireless communication networks.
%
%\item A novel graph representation learning model is proposed based on the constructed knowledge graph. In order to learn richer information about the graph structure, we add GAN to the graph encoder so that GCN can learn the node distribution and output a more accurate feature representation of the nodes.
%
%%\item On unsupervised node classification tasks on public datasets, the accuracy of the proposed model outperforms most baselines. On the knowledge graph of wireless communication networks, based on the experimental results, we propose a scheme for automatic classification of nodes in the graph and root cause tracking of abnormal data fields.
%\item The model shows excellent results on both the KG of wireless communication networks and other public graph datasets. Based on the experimental results, we propose a scheme for automatic classification of nodes in graphs and root cause tracking of anomalous data domains.
%
%\end{itemize}

\section{Construct a KG of wireless data of the core network}
Previous researches about intelligent wireless networks are based on simulation data. In contrast, our research adopts the real data collected via a big data platform, named as testbed for 5G/B5G intelligent network (TTIN). TTIN is a 5G/B5G field experimental network and is the world's first real-world data test platform for real-time wireless data collection, storage, analysis and intelligent closed-loop control \cite{33}. In this paper, we use the data containing all users in a network consisting of 12 trial cells from July to December 2021. Then, we construct a KG of the endogenous factors of the wireless data according to the entities and relations defined in the 3GPP Release 16 TS 23.502, 24.501, 38.413, etc, and the patent proposed by S. He et al.\cite{34}.
In what follows, we take the core network data to show how the KG is constructed.

\begin{table}[t]
	\caption{Categories of Entities in the KG.}
	\label{KGG}
	\begin{center}
		\renewcommand{\arraystretch}{1.5}
		\begin{tabular}{|c|c|c|}
			\hline
			\textbf{Category}   & \textbf{Number}  & \textbf{Example} \\ \hline
			Data field type   & 132   &    Msgflag  \\ \hline
			Procedure type      & 40   & Registration procedure   \\ \hline
			Statistical indicator    & 73    &  Regis request cnt     \\ \hline
			Algorithm indicator           & 2    &  Regis success rate   \\ \hline
			
		\end{tabular}
	\end{center}
\end{table}
\begin{table}[t]
	\caption{Categories of Relations in the KG.}
	\label{RE}
	\begin{center}
		\renewcommand{\arraystretch}{2}
		\begin{tabular}{|c|c|}
			\hline
			\textbf{Category}   & \textbf{Definition}  \\ \hline
			Procedure relation   & \makecell{The relation between procedure type\\ and data field type}   \\ \hline
			Condition relation      & \makecell{The relation between statistical indicator\\ and data field type}    \\ \hline
			Algorithm relation    & \makecell{The relation between statistical indicator\\ and algorithm indicator}     \\ \hline			
		\end{tabular}
	\end{center}
\end{table}
In general, KG is composed of entities and relations. Entities are objects or abstract concepts in the real world. Relations represent the relationships between entities.
According to the core network data fields and indicators as specified in the 3GPP Release 16 TS 23.502, 24.501, 38.413, etc, four categories of entities are established, including procedure type, data field type, statistical indicator, and algorithm indicator, as shown in Table \ref{KGG}. Among the four categories of entities, the procedure type refers to the procedures specified in China Mobile's 5G interface specification\cite{42} and owns a large proportion of data in TTIN, covering 40 entities, such as 1-Registration, 3-Service request, 4-Paging, etc.
The data field type refers to the data fields associated with the procedure types other than the location and basic device information. It includes 132 entities, such as msgflag, procedure status, request cause group, etc.
The statistical indicator refers to the indicators of a specific data field that can reflect the current bearer status of the network within a certain period. It includes 73 entities, such as regis request cnt, regis accept cnt, etc.
The algorithm indicator refers to indicators calculated by specific algorithms, such as regis success rate and regis failure rate.

\begin{figure}[t]
	\centerline{\includegraphics[width=3in]{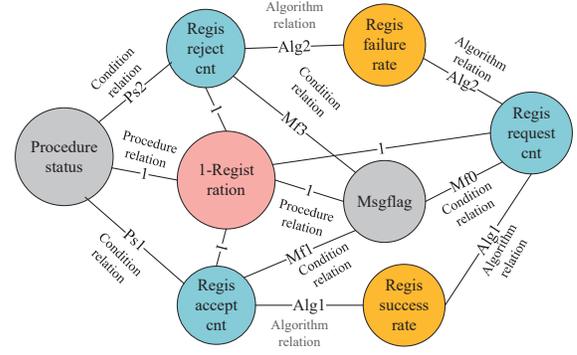}}
	\caption{A partial knowledge graph of the registration procedure.}
	\label{KG}
\end{figure}

Through the analysis of the collected wireless data, procedure relation, condition relation, and algorithm relation are defined, as shown in Table \ref{RE}.
The procedure relation describes which procedure type the data field type is associated with. For example, the rejected NSSAI number field is related to procedure 1-Registration.
The condition relation illustrates the value of a specific data field type associated with the statistical indicator. The algorithm relation describes the specific calculation method to convert statistical indicator into algorithm indicator. Fig.~\ref{KG} demonstrates a simple example of the KG, which includes four categories of entities and three categories of relations.

\begin{figure}[t]
	\centerline{\includegraphics[width=3.5in]{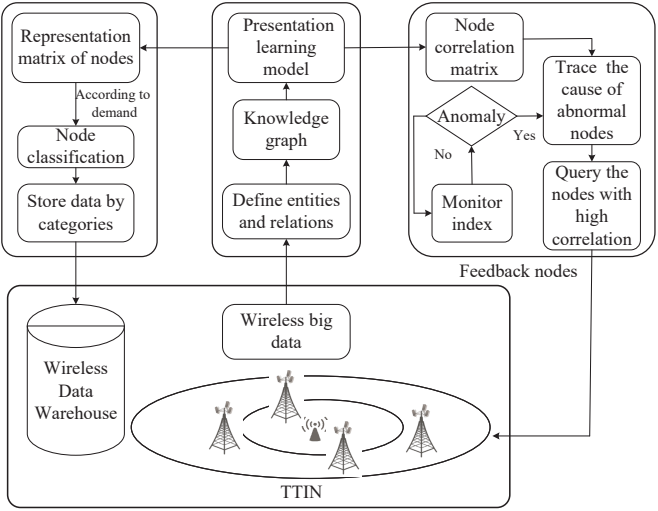}}
	\caption{Flowchart of applying the wireless data of the core network.}
	\label{procedure}
\end{figure}

After defining the entities and relations, a KG of wireless data of the core network can be constructed by regarding entities and relations as nodes and edges, respectively. Note that the classification of nodes is performed manually. Meanwhile, the number of categories and each category's data are fixed. Due to the dynamic characteristics of wireless networks, manually dividing and storing wireless data cannot meet the needs of data warehouses during different periods. To further understand and analyze the constructed KG in-depth, in Section \uppercase\expandafter{\romannumeral3}, we propose a representation learning model to obtain the representation vectors and the correlation matrix of nodes in the graph, then verify the model's performance on public datasets. In Section \uppercase\expandafter{\romannumeral4}, based on the KG constructed in Section \uppercase\expandafter{\romannumeral2}, we apply the model to automatically classify nodes with any number of categories and compare it with the manual classification results to verify the model's validity.
%If the constructed KG does not perform representation learning, the nodes and edges cannot be quantified, and the KG cannot be mined and reasoned. So in Section \uppercase\expandafter{\romannumeral3} we constructed a representation learning model and verified the model's performance on public datasets. Through the model, the representation vector of the node and the correlation matrix between the nodes can be obtained. These can be used to help build an intelligent data warehouse and trace the cause of network anomalies.
%In terms of classified storage of intelligent data warehouse, since the classification of graph nodes is performed manually, the number of categories and the data of each category are fixed. Due to the dynamic characteristics of wireless big data, manually dividing and storing data cannot meet the needs of data warehouses in different periods. Therefore, in Section \uppercase\expandafter{\romannumeral4}, we apply the model we built to classify nodes with any number of categories automatically and compare the model classification results with the manual classification result to verify the model's validity.
In addition, once the performance of communication networks is abnormal, the KG can be used to trace the network anomaly cause. In Section \uppercase\expandafter{\romannumeral4}, based on the KG constructed in Section \uppercase\expandafter{\romannumeral2}, we first verify the correctness of the correlation matrix output by the model in Section \uppercase\expandafter{\romannumeral3}, then explain how to trace the cause of the abnormal node through the correlation matrix. The overall framework of the paper is shown in Fig.~\ref{procedure}.

\section{Model for representation learning of KG}
%\subsection{General Framework}
In this section, we propose a new unsupervised neural network model for KRL of KG.
%The representation vector extracted from the model would integrate the information on the nodes itself and the neighboring nodes. Then the representation vectors are used in the downstream tasks, such as node classification, clustering, etc. The reconstructed adjacency matrix output by the model contain richer structural information, which can be used for reasoning tasks, correlation analysis between nodes, etc.
The representation vectors extracted from this model will integrate information from the nodes themselves and neighboring nodes, which can be used for downstream tasks such as classification and inference in Section \uppercase\expandafter{\romannumeral4}.
\begin{figure*}[t]
	\centerline{\includegraphics[scale=0.3]{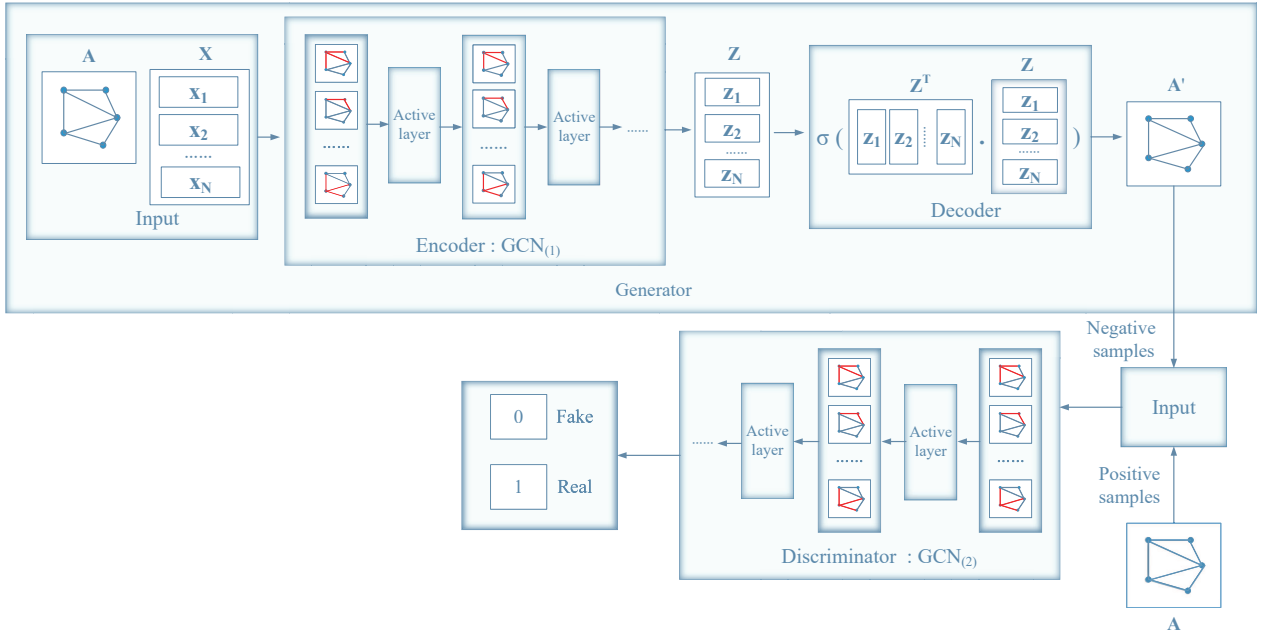}}
	\caption{The structure of the model for representation learning of KG.}
	\label{model}

\end{figure*}

\subsection{General Framework}
Let $G = (\mathbf{V}, \mathbf{E})$ be a given graph, where $\mathbf{E}$ is a set of edges and $\mathbf{V}$ is a set of $n$ nodes. The topological structure of graph $G$ can be represented by an adjacency matrix $\mathbf{A}$. The element $a_{ij}$ in the $i$-th row and $j$-th column of $\mathbf{A}$ indicates whether nodes are connected. $e_{ij}$ represents the edge between node $i$ and node $j$, when $e_{ij}\in \mathbf{E}$, $a_{ij}=1$; when $e_{ij}\notin \mathbf{E}$, $a_{ij}=0$. All diagonal elements $a_{ii}$ are set to 0. $N(v_{i})$ is defined as the set of nodes directly connecting to node $v_{i}$. The degree matrix $\mathbf D$ is a diagonal matrix with $ d_{ii}= \sum_{j=1}^n a_{ij}$.
Each node has its own attribute information. Feature matrix $\mathbf{X}\in R^{ n \times m}$ is used to represent the attribute information of each node, where $m$ is the dimension of the feature vector.

The KRL model under the GAN framework consists of a graph autoencoder as a generator and a discriminator based on GCN. The general framework is shown in Fig.~\ref{model}.
% it consists of two networks: generator $\mathcal{G}$ and discriminator $\mathcal{D}$.
The input of generator $\mathcal{G}$ is $\mathbf{A}$ and $\mathbf{X}$. The goal of generator $\mathcal{G}$ is to fit $p_{{\rm true}}(V|v_{i})$ and generate a reconstructed adjacency matrix $\mathbf{A'}$ similar to the real neighbors of the nodes to deceive the discriminator. $p_{{\rm true}}(V|v_{i})$ is the real node connectivity distribution, which reflects $v_{i}$’s connectivity preference distribution over all other nodes in $V$\cite{22}. It represents the preference of each node for connectivity with other nodes. Whereas the discriminator $\mathcal{D}$, on the contrary, takes $\mathbf{A}$ and $\mathbf{A}'$ as input, and tries to judge whether the adjacency of these nodes comes from the real graph or the graph generated by $\mathcal{G}$.

\subsection{Generator $\mathcal{G}$}
Generator $\mathcal{G}$ is realized by a graph autoencoder and includes two parts, i.e., an encoder and a decoder.

\paragraph{Encoder}
GCN is used to implement the encoder in this paper. The core of GCN is to learn a mapping function $f(\mathbf{X},\mathbf{A})$. This function can combine the neighbor node information and the node feature information to generate a new node representation matrix $\mathbf{Z}$. As a result, the multi-layer GCN has the following layered propagation rule in graph $G = (\mathbf{V}, \mathbf{E})$:
\begin{equation}
\label{GCN}
\mathbf{H}^{(l+1)}=\sigma(\tilde{\mathbf{D}}^{-1/2}\tilde{\mathbf{A}}\tilde{\mathbf{D}}^{-1/2}\mathbf{H}^{(l)}\mathbf{W}^{(l)}),
\end{equation}
where $\mathbf{H}^{(l)}$ represents the node representation of the $l$-th layer; $\mathbf{H}^{(0)}=\mathbf{X}$. There are $L$ layers in the GCN, and the output $\mathbf{Z}$ of the last layer of the encoder is the node representation matrix. $\tilde{\mathbf{A}}$ is the normalized adjacency matrix of graph $G$, $\tilde{\mathbf A}=\mathbf{A}+\mathbf{I}_{n}$, where $\mathbf{I}_{n}$ is identity matrix. The degree matrix $\tilde{\mathbf D}$ is a diagonal matrix with $\tilde{d}_{ii}= \sum_{j=1}^n \tilde{a}_{ij}$. $\mathbf W^{(l)}$ is a trainable parameter matrix in the neural networks. $\sigma(.)$ represents the activation function. It should be pointed out that the two layer GCN based on (\ref{GCN}) is proposed by Kipf et al. \cite{19}.

\paragraph{Decoder}
The object of the decoder is to reconstruct the adjacency matrix. Since the encoder compresses the data into a low-dimensional, compact, and continuous feature space, the decoder needs to reconstruct the adjacency matrix through the node representation matrix $\textbf Z$. We define the decoder as follows:
%We define the decoder as follows (\ref{loss1}) by adopting the method of Kipf et al.\cite{20}.
\begin{equation}
\label{loss1}
 \hat{\textbf{A}} = \sigma(\textbf Z^{\text{T}}\textbf Z).
\end{equation}
$\hat{\textbf{A}}$ is the correlation matrix. The value of each element in $ \hat{\textbf{A}}$ is between 0 and 1, it can be regarded as the probability or closeness of the connection between nodes. The parameters of the GCN in the encoder can be optimized by minimizing the reconstruction error of the model:
\begin{equation}
\label{loss}
\mathcal L = -\dfrac{1}{n^{2}}\sum_{i=1}^{n}\sum_{j=1}^{n}\hat{a}_{ij}\log(a_{ij}),
\end{equation}
After training, the node representation matrix $\textbf Z$ generated by GCN can characterize the graph structure more accurately. In addition to using the loss function in (\ref{loss}), the loss function of GAN can also be used to train the encoder to enhance its generation ability, which is defined in subsection C.

\subsection{Discriminator $\mathcal{D}$}
The purpose of the discriminator $\mathcal{D}$ is to distinguish whether the input is $\textbf A$ (positive) or $\textbf A'$ (negative), where $\textbf A'=\mathcal{G}(\textbf X,\textbf A)$. $\mathcal{D}(\textbf A)$ and $\mathcal{D}(\mathcal{G}(\textbf X,\textbf A))$ represent the scoring of the real adjacency matrix and the reconstructed adjacency matrix by the discriminator $\mathcal{D}$, respectively. In fact, generator $\mathcal{G}$ and discriminator $\mathcal{D}$ are playing the following two-player minimax game with the loss function $\mathcal L_{gan}$:
\begin{equation}
\label{GAN}
\begin{aligned}
\min\limits_{\theta_{\mathcal G}}\max\limits_{\theta_{D}}\mathcal L_{gan}=& E_{\textbf A\sim p_{ture}}[\log\mathcal{D}(\textbf A)]+\\
& E_{\textbf{A}'\sim p_{_\mathcal G}}[\log(1-\mathcal{D}(\mathcal{G}(\textbf X, \textbf A)))],
\end{aligned}
\end{equation}
The parameter optimization of generator $\mathcal{G}$ and discriminator $\mathcal{D}$ can be realized by solving the problem (\ref{GAN}). In each iteration, the parameters of the generator are firstly optimized through (\ref{loss}). Then the reconstructed adjacency matrix and real adjacency matrix are fed into the discriminator, and finally optimize the parameters of the discriminator and generator through (\ref{GAN}).
\subsection{Model Evaluation}
In this subsection, in order to demonstrate the superiority of the performance of the proposed model, we compare it with the results of other representation models on the classical task of representation learning, namely unsupervised node classification. We use Cora, Citeseer, and PubMed, the most commonly used public datasets in the field of graph neural networks, to verify the superiority of the proposed model compared to other representation models.
%These three datasets have 2704, 3327, 19717 nodes and 5429, 4732, 44338 edges respectively.

%\paragraph{Datasets}
%The specific information on the three datasets, i.e., Cora, Citeseer and PubMed, is summarized in Table \ref{data set}. According to disciplines, the three datasets are divided into 7, 6, and 3 categories, respectively.
%%In the task of unsupervised node classification, the category labels are not used to train the model, but only as the test results.
%\begin{table}[t]
%	\caption{Information on Datasets.}
%	\label{data set}
%	\renewcommand{\arraystretch}{1.2}
%	\begin{center}
%		\begin{tabular}{|c|c|c|c|c|}
%			\hline
%			\textbf{Dataset} & \textbf{Type}    & \textbf{Nodes} & \textbf{Edges} & \textbf{Classes} \\ \hline
%			Cora       & Citation network & 2708         &5429         & 7                \\ \hline
%			Citeseer             & Citation network & 3327           & 4732           & 6                 \\ \hline
%			PubMed           & Citation network & 19717          & 44338          & 3                \\ \hline
%		\end{tabular}
%	\end{center}
%\end{table}
\paragraph{Baseline}
We investigate the methods based on graph embedding, including DeepWalk \cite{25}, node2vec \cite{26}, LINE \cite{27} and GraRep \cite{28}, as well as compared the methods based on unsupervised graph neural network, including VGAE \cite{20}, ARVGE \cite{29}, DGI \cite{30}, OT-GCN and OT-GAT\cite{31}. In particular, VGAE is a graph autoencoder that does not use GAN, and comparing our method with it shows the enhancement of the encoder's generation capability by adding GAN.

\paragraph{Node Classification}
For the unsupervised node classification task, we feed adjacency matrix $\textbf A$ and the feature matrix $\textbf X$ into the model. After obtaining the node representation matrix $\textbf Z$, K-means clustering is performed on $\textbf Z$ with accuracy (ACC) being the evaluation indicator. The comparison results with the baseline are shown in Table ~\ref{result1}. In Cora dataset, our method achieves an accuracy of 0.721, the highest among all methods. In Citeseer dataset, our method has the same accuracy as OT-GAT. In PubMed dataset, the accuracy of our method is higher than OT-GCN, but slightly lower than OT-GAT.

\begin{table}[htbp]
	\caption{Performance (\%) on the Node Classification Task.}
	\label{result1}
	\begin{center}
		\renewcommand{\arraystretch}{1.2}
		\begin{tabular}{llll}
			\hline
%			\multirow{2}*{Methods} & \multicolumn{2}{l}{\textbf{Cora}} & \multicolumn{2}{l}{\textbf{Citeseer}} & \multicolumn{2}{l}{\textbf{PubMed}} \\ \hline
			\textbf{Methods} & \textbf{Cora} & \textbf{Citeseer} & \textbf{PubMed} \\ \hline
%			& \textbf{ACC}        & \textbf{ACC}          & \textbf{ACC}         \\ \hline
			\textbf{DeepWalk}                 & 0.456                      & 0.362                        & 0.649                        \\
			\textbf{Node2Vec}                 & 0.563                      & 0.408                        & 0.656                       \\
			\textbf{LINE}                     & 0.308                      & 0.250                        & 0.431                        \\
			\textbf{GraRep}                   & 0.483                      & 0.312                        & 0.544                        \\ \cline{1-4}
%			\textbf{GAE}                      & 91.02                      & 89.54                        & 96.40                      \\
			\textbf{VGAE}                     & 0.571                      & 0.535                        & 0.586                        \\
%			\textbf{ARGE}                     & 92.43                      & 91.93                        & 96.81                        \\
			\textbf{ARVGE}                    & 0.641                      & 0.435                        & 0.588                        \\
			\textbf{DGI}                      & 0.635                      & 0.675                        & 0.641                        \\
			\textbf{OT-GCN}                   & 0.646                      & 0.689                        & 0.664                        \\
			\textbf{OT-GAT}                   & 0.667                      & 0.695                        & \textbf{0.673}                        \\			
			\textbf{Our method}                & \textbf{0.721}             & \textbf{0.695}               & 0.667          \\ \hline
		\end{tabular}
	\end{center}
\end{table}

\section{Applications on the KG of wireless core network data}
In this section, we analyze the KG of wireless core network data constructed in Section \uppercase\expandafter{\romannumeral2} based on the output of the proposed model. The experimental simulations are carried out based on Python Tensorflow framework.
Two main tasks will be accomplished: 1) Node classification. The application of the node classification is to build an on-demand data warehouse theme library. The specific method is to classify the nodes according to the node representation matrix $\textbf Z$ generated by the model. 2) Correlation analysis. The application of the correlation analysis is to assist in the cause tracing of anomalies in the wireless communication network. If a node is abnormal, a series of nodes that have the most significant impact on this abnormal node would be queried according to the correlation matrix $\hat{ \textbf A}$ output by the model.

\subsection{Node Classification}
In this subsection, we use the proposed model in Section \uppercase\expandafter{\romannumeral3} to classify nodes of the KG constructed in Section \uppercase\expandafter{\romannumeral2}. According to the topology of the graph, we build an adjacency matrix of size $247\times247$. As a baseline, the nodes in the graph are manually divided into four categories, which are listed in Table \ref{KGG}.

\begin{figure}[t]
	\centerline{\includegraphics[width=3in]{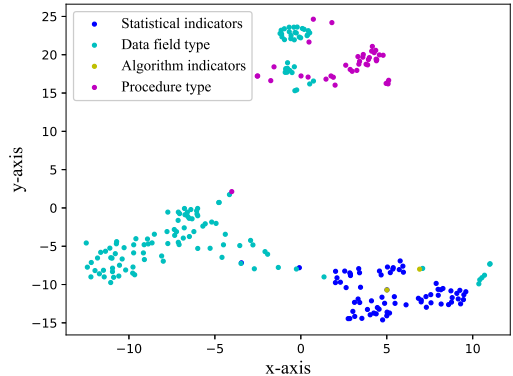}}
	\caption{Visualization of node classification by KG (4 categories).}
	\label{embedding}
\end{figure}

Under unsupervised conditions, i.e., without label information, we investigate the difference between the results of model classification and manual classification. Specifically, we feed the adjacency matrix of the KG into the model and then perform K-means clustering algorithm on the node representation matrix of the model. Finally, the cross-entropy between the model classification results and the manual classification results is calculated to measure the classification accuracy. The classification accuracy increases with more iterations and eventually converges at 80.1\%, i.e., the coincidence rate of the classification results with expert knowledge is 80.1\%. Fig.~\ref{embedding} displays the node classification result of KG by the proposed model with the t-SNE visualization method. The high classification accuracy illustrates that the node representations learned by the model contain sufficient node information and graph structure information used for automatic node classification. So the representation vector of each node obtained through the model, combined with some common clustering methods, can realize the construction of different data warehouses on demand.
%Fig.~\ref{embedding} displays the node classification result of KG with the t-SNE visualization method\cite{32}.

%Following the method above, we classify the nodes into 6, 8, 10 and 12 classes. Fig. \ref{class} shows the classification results using the t-SNE visualization method \cite{32} when the nodes are divided according to the requirements.

%According to different requirements, our constructed model combined with K-means can generate embedding vectors for various categories, as shown in Fig. ~\ref{class}, which sets the category to 6, 8, 10, and 12.

%\begin{figure}[t]
%	\centerline{\includegraphics[width=3.5in]{Visualize_many.eps}}
%	\caption{Visualization of node classification of KG (6, 8, 10 and 12 categories).}
%	\label{class}
%\end{figure}

\subsection{Correlation Analysis}
After the node classification task, we carry out the correlation analysis between nodes of the constructed KG. This subsection first evaluates the correlation matrix's correctness and then introduces how the correlation matrix can be used to trace the cause of abnormal nodes.
The edge set is divided into training and test sets for the same datasets mentioned above, accounting for 90\% and 10\%, respectively. The model calculates and outputs the correlation matrix according to (\ref{loss1}).
Based on the test set and the corresponding correlation matrix, the prediction average accuracy (AP) and the area under the curve (AUC) of the model can be calculated.

As the number of iterations increases, the test set's AP gradually increases and finally converges to 0.929, while AUC converges to 0.857.
Then, we adjusted the ratio of the training set and to test set, and compared the prediction results when the test set proportion was set to 5\% ,10\%, 20\%, 30\%, 50\%. As shown in Tabel \ref{yuzhi}, the experimental results show that the correlation matrix obtained by the proposed model has high correctness.

\begin{table}[t]
	\caption{Model Performance (\%) of Relation Prediction.}
	\label{yuzhi}
	\renewcommand{\arraystretch}{1.2}
	\begin{center}
		\begin{tabular}{|c|c|c|c|c|c|}
			\hline
			\textbf{Test set ratio} & 5\%   & 10\%  & 20\%  & 30\%  & 50\%  \\ \hline
			\textbf{AP}             & 0.942 & 0.929 & 0.923 & 0.930 & 0.899 \\ \hline
			\textbf{AUC}            & 0.884 & 0.857 & 0.847 & 0.861 & 0.798 \\ \hline
		\end{tabular}
	\end{center}
\end{table}

\begin{figure}[t]
	\centerline{\includegraphics[width=3in]{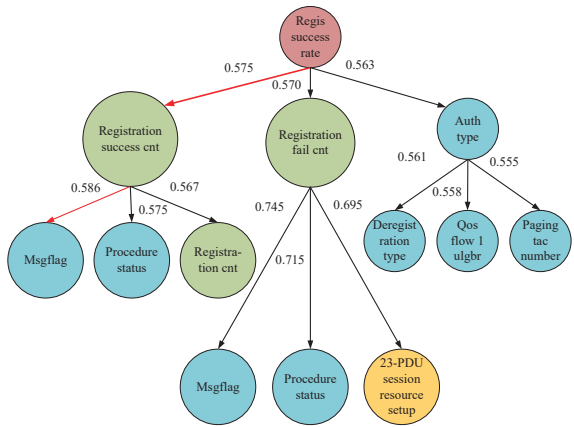}}
	\caption{The two-level association tree of regis success rate.}
	\label{shiyitu}
\end{figure}

%\balance
%With the help of the correlation matrix, a series of nodes that influence the abnormal node in the KG can be quickly found out. In other words, the cause of network anomaly can be traced.
With the help of the correlation matrix, the cause of the network anomaly can be traced. Construct an association tree using the correlation matrix, and then trace the cause of network anomalies by the association tree. The specific calculation steps are summarized in Algorithm 1.
\begin{algorithm}[htb]
	\caption{ Tracing the cause of network anomaly.}
	\label{alg:Framwork}
	\begin{algorithmic}[1] %这个1 表示每一行都显示数字
		\REQUIRE ~~\\ %算法的输入参数：Input
		The abnormal node in the KG of wireless data of the core network, $v^*$;\\
		The number of levels of the association tree, $l$;\\
		The number of degrees of each node in the tree, $m$;\\
		The correlation matrix, $\hat{\textbf{A}}$
		\ENSURE ~~\\ %算法的输出：Output
		 A series of nodes that influence abnormal node, $\textbf V^*$;
		\STATE Initialize a tree $\textbf T$ with the root node $v^*$, and use it as the 0-th level of the tree;
		\STATE Place $v^*$ in the 0-th level node set $\textbf V_0$;		
		\FOR{each $i \in [1,l]$}
		\FORALL{$v_c \in \textbf V_{i-1}$}
		\STATE Build the $i$-th level of association tree using $m$ nodes with the highest correlation with $v_{c}$ in matrix $\hat{\textbf A}$ and add these nodes to $\textbf V_{i}$;\
		\STATE Take the correlation between nodes as the weight of the corresponding edge;\
		\ENDFOR
		\ENDFOR
		\STATE Finish building the association tree;
		\STATE Initialize the $v^*$ as $v_0^*$ ;
		\FOR{each $i \in [1,l]$}
		\STATE In the $i$-th layer, find node $v_i^*$, which is connected node of $v_{i-1}^*$ and with the highest correlation with $v_{i-1}^*$;\
		\STATE Place $v_i^*$ in $\textbf V^*$
		\ENDFOR
		\RETURN $\textbf V^*$; %算法的返回值
	\end{algorithmic}
\end{algorithm}
As shown in Fig. \ref{shiyitu}, the regis success rate is taken as the abnormal node, and the most straightforward two-level association tree is used as an example.
In the first level, the edge weight from the root node to registration success cnt is 0.575, which is greater than 0.570 of registration fail cnt and 0.563 of auth type, which means that registration success cnt has the highest correlation with the abnormal node of registration success rate detected by the system.
Similarly, in the second level, msgflag has the highest correlation with registration success cnt. This means that the msgflag data field is most likely to be the cause of abnormalitiy of the regis success rate. If it is necessary to get the deeper reason, the operation and maintenance (O\&M) personnel can deepen the level of the association tree according to the above algorithm, and then continue to search for corresponding nodes. In this way, O\&M personnel can search for nodes associated with abnormal nodes step by step, thereby tracing the cause of the network anomaly.

\balance

\section{Conclusion}
We have established the KG of wireless data of the core network for the first time and proposed a novel model for graph representation learning. Then, we perform node classification and relation prediction simulations on the KG of wireless data of the core network. These studies are based on deep learning rather than traditional algorithms, which can help build intelligent wireless communication networks data warehouses and infer to trace the cause of the anomaly in wireless networks.

\section{ACKNOWLEDGEMENT}
This work was supported in part by National Natural Science Foundation of China under Grants 62171474 and  61720106003, in part by the National Key Research and Development Program under Grant 2021YFB2900300, in part by the open research fund of National Mobile Communications Research Laboratory Southeast University under Grants No. 2022D03, in part by OPPO research fund under Grants No. CN05202112160224, in part by the Major Key Project of PengCheng Laboratory under Grants PCL2021A01-2.
%\balance
\bibliographystyle{IEEEtran}
\bibliography{reference}

\begin{thebibliography}{10}
\providecommand{\url}[1]{#1}
\csname url@samestyle\endcsname
\providecommand{\newblock}{\relax}
\providecommand{\bibinfo}[2]{#2}
\providecommand{\BIBentrySTDinterwordspacing}{\spaceskip=0pt\relax}
\providecommand{\BIBentryALTinterwordstretchfactor}{4}
\providecommand{\BIBentryALTinterwordspacing}{\spaceskip=\fontdimen2\font plus
\BIBentryALTinterwordstretchfactor\fontdimen3\font minus
  \fontdimen4\font\relax}
\providecommand{\BIBforeignlanguage}[2]{{%
\expandafter\ifx\csname l@#1\endcsname\relax
\typeout{** WARNING: IEEEtran.bst: No hyphenation pattern has been}%
\typeout{** loaded for the language `#1'. Using the pattern for}%
\typeout{** the default language instead.}%
\else
\language=\csname l@#1\endcsname
\fi
#2}}
\providecommand{\BIBdecl}{\relax}
\BIBdecl

\bibitem{37}
P.~Dong, H.~Zhang, G.~Y. Li\emph{,~et~al.}, ``Deep cnn-based channel estimation
  for mmwave massive mimo systems,'' \emph{IEEE J. Sel. Topics Signal Proc.},
  vol.~13, no.~5, pp. 989--1000, Jul. 2019.

\bibitem{38}
Y.~He, C.~Liang, F.~R. Yu\emph{,~et~al.}, ``Optimization of cache-enabled
  opportunistic interference alignment wireless networks: A big data deep
  reinforcement learning approach,'' in \emph{Proc. IEEE ICC}, Jul. 2017, pp.
  1--6.

\bibitem{41}
S.~Rezaie, C.~N. Manchón, and E.~de~Carvalho, ``Location- and
  orientation-aided millimeter wave beam selection using deep learning,'' in
  \emph{Proc. IEEE ICC}, Jul. 2020, pp. 1--6.

\bibitem{35}
S.~He, Y.~Zhang, J.~Wang\emph{,~et~al.}, ``A survey of millimeter-wave
  communication: Physical-layer technology specifications and enabling
  transmission technologies,'' \emph{Proc. IEEE}, vol. 109, no.~10, pp.
  1666--1705, Oct. 2021.

\bibitem{1}
S.~Ji, S.~Pan, E.~Cambria\emph{,~et~al.}, ``A survey on knowledge graphs:
  Representation, acquisition, and applications,'' \emph{IEEE Trans. Neur. Net.
  Lear.}, pp. 1--21, Apr. 2021.

\bibitem{18}
I.~Goodfellow, J.~Pouget-Abadie, M.~Mirza\emph{,~et~al.}, ``Generative
  adversarial nets,'' vol.~27, Dec. 2014.

\bibitem{33}
Y.~Huang, S.~Liu, C.~Zhang\emph{,~et~al.}, ``True-data testbed for {5G/B5G}
  intelligent network,'' \emph{Intell. Converged Networks}, vol.~2, no.~2, pp.
  133--149, Jun. 2021.

\bibitem{34}
S.~He, L.~Wang, X.~Zhan\emph{,~et~al.}, ``Methods, systems, equipment and media
  for constructing and analyzing wireless network protocol knowledge graphs(in
  chinese),'' China Patent CN112\,714\,032B, Jul. 2, 2021.

\bibitem{42}
\emph{Technical Specification of Deep Packet Inspection Equipment for CMCC
  (Signalling Collection Server Equipment Part)}, China Mobile Communications
  Group China Mobile Communication Enterprise Standard 1.0.0, Mar. 2020.

\bibitem{22}
H.~Wang, J.~Wang, J.~Wang\emph{,~et~al.}, ``Learning graph representation with
  generative adversarial nets,'' \emph{IEEE Trans. Knowl. Data En.}, vol.~33,
  no.~8, pp. 3090--3103, Aug. 2021.

\bibitem{19}
T.~N. Kipf and M.~Welling, ``Semi-supervised classification with graph
  convolutional networks,'' \emph{arXiv preprint arXiv:1609.02907}, Sep. 2016.

\bibitem{25}
B.~Perozzi, R.~Al-Rfou, and S.~Skiena, ``Deepwalk: Online learning of social
  representations,'' in \emph{Proc. ACM SIGKDD}, Mar. 2014, pp. 701--710.

\bibitem{26}
A.~Grover and J.~Leskovec, ``node2vec: Scalable feature learning for
  networks,'' in \emph{Proc. ACM SIGKDD}, Aug. 2016, pp. 855--864.

\bibitem{27}
J.~Tang, M.~Qu, M.~Wang\emph{,~et~al.}, ``Line: Large-scale information network
  embedding,'' in \emph{Proc. WWW}, May 2015, pp. 1067--1077.

\bibitem{28}
S.~Cao, W.~Lu, and Q.~Xu, ``Grarep: Learning graph representations with global
  structural information,'' in \emph{Proc. ACM CIKM}, Oct. 2015, pp. 891--900.

\bibitem{20}
T.~N. Kipf and M.~Welling, ``Variational graph auto-encoders,'' \emph{arXiv
  preprint arXiv:1611.07308}, Nov. 2016.

\bibitem{29}
S.~Pan, R.~Hu, G.~Long\emph{,~et~al.}, ``Adversarially regularized graph
  autoencoder for graph embedding,'' in \emph{Proc. IJCAI}, Jul. 2018, pp.
  2609--2615.

\bibitem{30}
P.~Veli{\v{c}}kovi{\'c}, W.~Fedus, W.~L. Hamilton\emph{,~et~al.}, ``Deep graph
  infomax,'' \emph{arXiv preprint arXiv:1809.10341}, Sep. 2018.

\bibitem{31}
L.~Yang, J.~Gu, C.~Wang\emph{,~et~al.}, ``Toward unsupervised graph neural
  network: Interactive clustering and embedding via optimal transport,'' in
  \emph{Proc. IEEE ICDM}, Nov. 2020, pp. 1358--1363.

\end{thebibliography}
\end{document}